\documentclass[sigconf]{acmart}
\usepackage{comment}
\usepackage{color, colortbl}
\definecolor{red}{rgb}{0.9,0.7,0.7}
\definecolor{green}{rgb}{0.7,0.9,0.7}
\definecolor{magenta}{rgb}{1,0,1}
\definecolor{blue}{rgb}{0.7,0.7,0.9}
\definecolor{grey}{rgb}{0.9,0.9,0.9}

\newcommand{\spheading}[2][3em]{% \spheading[<width>]{<stuff>}
  \rotatebox{90}{\parbox{#1}{\raggedright #2}}}

\def\BibTeX{{\rm B\kern-.05em{\sc i\kern-.025em b}\kern-.08emT\kern-.1667em\lower.7ex\hbox{E}\kern-.125emX}}

\newcommand{\etal}{\textit{et al}. }

\begin{document}

%
% The "title" command has an optional parameter, allowing the author to define a "short title" to be used in page headers.
\title{RIT-Eyes: Rendering of near-eye images for eye-tracking applications}

\author{Nitinraj Nair}
\email{nrn2741@rit.edu}
\affiliation{%
  \institution{Rochester Institute of Technology}
  \city{Rochester}
  \state{NY}
  \country{USA}
  \postcode{14623}
}
\author{Rakshit Kothari}
\email{rsk3900@rit.edu}
\orcid{0002-4318-3068}
\affiliation{%
  \institution{Rochester Institute of Technology}
  \city{Rochester}
  \state{NY}
  \country{USA}
} 

\author{Aayush K. Chaudhary}
\email{akc5959@rit.edu}
\affiliation{%
  \institution{Rochester Institute of Technology}
  \city{Rochester}
  \state{NY}
  \country{USA}}
%\email{akc5959@rit.edu}

 \author{Zhizhuo Yang}
 \email{zy8981@rit.edu}
 \affiliation{%
   \institution{Rochester Institute of Technology}
  \city{Rochester}
   \state{New York}
   \country{USA}
 }
%\email{akc5959@rit.edu}

\author{Gabriel J. Diaz}
\email{gabriel.diaz@rit.edu}
\affiliation{%
  \institution{Rochester Institute of Technology}
  \city{Rochester}
  \state{NY}
  \country{USA}}

\author{Jeff B. Pelz}
\email{jeff.pelz@cis.rit.edu}
\affiliation{%
  \institution{Rochester Institute of Technology}
  \city{Rochester}
  \state{NY}
  \country{USA}}

\author{Reynold J. Bailey}
\email{rjb@cs.rit.edu}
\affiliation{%
  \institution{Rochester Institute of Technology}
  \city{Rochester}
  \state{NY}
  \country{USA}}

%
% By default, the full list of authors will be used in the page headers. Often, this list is too long, and will overlap
% other information printed in the page headers. This command allows the author to define a more concise list
% of authors' names for this purpose.

%
% The abstract is a short summary of the work to be presented in the article.
\begin{abstract}
Deep neural networks for video-based eye tracking have demonstrated resilience to noisy environments, stray reflections, and low resolution. However, to train these networks, a large number of manually annotated images are required. To alleviate the cumbersome process of manual labeling, computer graphics rendering is employed to automatically generate a large corpus of annotated eye images under various conditions. In this work, we introduce a synthetic eye image generation platform that improves upon previous work by adding features such as an active deformable iris, an aspherical cornea, retinal retro-reflection, gaze-coordinated eye-lid deformations, and blinks. To demonstrate the utility of our platform, we render images reflecting the represented
 gaze distributions inherent in two publicly available datasets, NVGaze and OpenEDS. We also report on the performance of two semantic segmentation architectures (SegNet and RITnet) trained on rendered images and tested on the original datasets.
% while testing on the original datasets.
\end{abstract}


% ZY ============   REPLACE THE CODE BELOW    ==============


%
% Keywords. The author(s) should pick words that accurately describe the work being
% presented. Separate the keywords with commas.

% teaser image goes here
\begin{teaserfigure}
  \centering
  \includegraphics[width=1\textwidth]{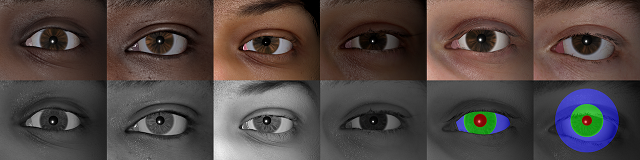}
  \caption{Sample images generated using our  synthetic eye image generation platform. RGB images (top). Corresponding simulated infrared images (bottom). Pixels assigned a label through the process of semantic segmentation are shown on the bottom rightmost images with and without skin (to illustrate placement of the eyeball). Red = pupil, green = iris, blue = sclera.}
  \label{fig:head_models}
\end{teaserfigure}

%
% This command processes the author and affiliation and title information and builds
% the first part of the formatted document.
\maketitle

% \begin{figure*}
% \centering
% \includegraphics[width=17.7cm, height=4.8cm]{images/all-face2.png}
%   \caption{Sample images generated using our  synthetic eye image generation platform. RGB images (top). Corresponding infrared images (bottom).}
% \label{fig:head_models}
% \end{figure*}

\section{Introduction and Related Work}
Modern video-based eye-trackers use infrared cameras to monitor movements of the eyes in order to gather information about the visual behavior and perceptual strategies of individuals engaged in various tasks. Eye-trackers have traditionally been mounted to computer screens or worn directly on the head for mobile applications and are increasingly being embedded in head-mounted-displays to support rendering and interaction in virtual and augmented reality applications. Contemporary algorithms for estimating gaze direction rely heavily on the segmentation of specific regions of interest in the eye images, such as the pupil (see Figure~\ref{fig:head_models} bottom right). These features are then used to estimate the causal 3D geometry, in the form of a 3D model of the spherical eye in camera space~\cite{Swirski2013AFitting}. Segmentation is complicated by the presence of corrective lenses or by reflections of the surrounding environment upon intervening physiology, such as the cornea and the tear layer. Convolutional neural networks (CNNs) have presented a promising new approach for the annotation of eye images even in these challenging conditions~\cite{Fuhl2017PupilNetDetection, Zhang2015Appearance-basedWild, Yiu2019DeepVOG:Learning, Kim2019NVGaze:Estimation, Linden2018LearningTracking, Vera-Olmos2018DeepEye:Environments}. However promising, the use of CNNs trained through supervised learning requires a large corpus of annotated eye images. Manual annotation is time-intensive and, although these datasets exist, they are susceptible to errors introduced during human annotation, and only a few exist that include manually segmented image features other than the pupil center which are necessary for building an accurate 3D model of the eye. For a comprehensive list of existing datasets, please refer to Garbin \etal~\cite{Garbin2019OpenEDS:Dataset}. Although existing dataset images may include reflections of the surrounding environment, their inclusion has been unsystematic, and their contributions to the robustness of segmentation remain unclear.

%\begin{figure}[t]
%\begin{center}
%\includegraphics[width=8.3cm, %height=4.2cm]{images/all_eyes.png}
%\end{center}
   %\caption{Example IR images from the dataset. The bottom row from left to right shows the effects of the aspherical cornea, tear film, pupil size variation(1-4mm radius), and the presence of reflective glass. For comparison, the top row shows the result with the corresponding feature disabled. }
%\label{fig:all_effects}
%\end{figure}

To alleviate these limitations, solutions proposed by Bohme et al.~\cite{Bohme2008ATrackers}, Swirski et al.~\cite{Swirski2014RenderingEvaluation}, Wood et al.~\cite{Wood2016LearningImages, Wood2015RenderingEstimation}, and Kim et al.~\cite{Kim2019NVGaze:Estimation} involve rendering near-eye images in which the location of image features is known, circumventing the need for manual annotation. In this work, we present a new dataset (see example images in Figure~\ref{fig:head_models}), which builds on the success of our predecessors. Similar to previous efforts, we use a 3D model of the human eye to render 2D imagery similar to what is captured by an eye-tracker. Our synthetic eye image generation platform introduces several improvements  including an accurate aspherical corneal model, a deformable iris, the lacrimal caruncle (the small pink nodule located at the inner corner of the eye), gaze-coordinated blinks, and a retina with retro-reflective properties which can aid in developing `bright pupil' solutions~\cite{Morimoto2005EyeApplications} for head-mounted or remote eye trackers (see section 3 and Table~\ref{tbl:synthetic_datasets} for a comprehensive list of improvements).

The real evaluation of the effectiveness of any synthetic dataset lies in its ability to be leveraged in real-world problems. Although initial efforts demonstrate that CNNs trained on artificial stimuli for semantic segmentation can generalize to true imagery~\cite{Kim2019NVGaze:Estimation, Park2018LearningSettings, Park2018DeepEstimation}, these initial tests are limited to specific applications. Kim et al.~\cite{Kim2019NVGaze:Estimation} showed that despite their best efforts to model realistic distributions of natural eye imagery in a virtual reality headset, training on synthetic eye images while testing on real data resulted in a 3.1$^\circ$ accuracy error on average, which is $\sim 1^\circ$ higher than training on real imagery. Park et al. utilized the UnityEyes dataset~\cite{Wood2015RenderingEstimation} to train a stacked hourglass architecture~\cite{Newell2016StackedEstimation} to detect landmarks in real-world eye imagery. By augmenting the synthetic data with a few real-world images, they observed an improvement in performance~\cite{Park2019Few-shotEstimation}.
Together these studies suggest that the underlying distribution within existing synthetic datasets cannot capture the variability observed in real world eye imagery. While techniques such as few shot learning~\cite{Park2019Few-shotEstimation} and joint learning/un-learning~\cite{Alvi2019TurningEmbeddings} may help combat these issues, an inherently better training set distribution should aid in improving the performance of convolutional networks.

% \aayush{I feel this and last paragraph has repetitive texts..I would suggest to place this paragraph as Our major contributions are :1,2,3,...}

Here, we present a novel synthetic eye image generation platform and test its utility in the development of CNNs for semantic segmentation. This test involves using our framework to render three synthetic datasets: two that approximate the eye/camera/emitter position properties of publicly available datasets, one synthetic - NVGaze~\cite{Kim2019NVGaze:Estimation}, and one of real eye imagery - OpenEDS~\cite{Garbin2019OpenEDS:Dataset}. The third dataset approximates the eye/camera/emitter properties of the Pupil Labs Core wearable eye tracker~\cite{Kassner2014Pupil:Interaction} and is referred to as S-General. Renderings which mimic NVGaze-synthetic and OpenEDS will be referred to as S-NVGaze and S-OpenEDS respectively. These datasets enable us to test the generalizability of our rendering pipeline. For example, if two CNNs trained on S-NVGaze and S-OpenEDS respectively exhibit little or no difference in performance when tested on an external image, we can conclude that properties differentiating S-NVgaze and S-OpenEDS (namely the camera orientation, and placement and number of emitters) do not contribute to the semantic understanding of different eye regions in the external image.
% \rk{For example, if CNNs trained on our synthetic datasets exhibit little to no difference when tested on NVGaze, we can conclude that properties differentiating S-NVgaze and S-OpenEDS do not contribute to the semantic understanding of different eye regions.}

% \rb{RK, A BIT VAGUE/AMBIGUOUS - ATTEMPTING TO CLARIFY, STILL NEEDS WORK: For example, if two CNNs trained on S-NVGaze and S-OpenEDS respectively exhibit little or no difference in performance when tested on \rk{any external image} \sout{NVGaze}, we can conclude that properties differentiating S-NVgaze and S-OpenEDS (namely the camera orientation, placement and number of emitters) do not contribute to the semantic understanding of different eye regions \rk{in the said external image}.} 

This work records the performance of two semantic segmentation models with varying complexity (SegNet~\cite{Badrinarayanan2017SegNet:Segmentation} and RITnet~\cite{Chaudhary2019RITnet:Tracking}) trained on S-OpenEDS and S-NVgaze followed by testing on all available datasets (S-NVGaze, S-OpenEDS, S-General, NVGaze and OpenEDS). Network performance is used to identify the factors which limit our pipeline from generalizing across varying eye appearances and camera positions.

\begin{table}[t]
\centering
 %\rotatebox{90}{\textbf{Unity Eyes}}
\begin{tabular}{c|c|c|c|c|c} 
%\hline
 %&  \rotatebox{90}{\textbf{Unity Eyes}} & \rotatebox{90}{\textbf{Synth Eyes}} & \rotatebox{90}{\textbf{NVGaze}}    &\rotatebox{90}{\textbf{Swirski}}   & \rotatebox{90}{\textbf{Ours}}   \\ 
 
  &  \spheading{\textbf{Unity Eyes}} & \spheading{\textbf{Synth Eyes}} & \spheading{\textbf{NVGaze}}    &\spheading{\textbf{Swirski}} & \spheading{\textbf{Ours}}   \\ 
\hline
\textbf{Aspherical cornea}   & {$\times$}                                                 & {$\times$}                                                  & {$\times$}  & {$\times$}  &      \checkmark         \\ [1ex]
\hline
\textbf{Retroreflection}   & {$\times$}                                                 & {$\times$}                                                  & {$\times$}  & {$\times$}  &    \checkmark           \\ [1.5ex]
\hline
\textbf{Segmentation mask} & {$\times$}                                                 & {$\times$}                                                  &    \checkmark         & {$\times$}  &      \checkmark         \\ [1ex]
\hline
\textbf{Infrared rendering} & {$\times$}  & {$\times$}   &       \checkmark      &     \checkmark        &       \checkmark        \\ [1ex]
\hline
\textbf{RGB rendering}  &  \checkmark   &     \checkmark                          & {$\times$}  & {$\times$}  &  \checkmark             \\ [1ex]
\hline
\textbf{Reflective eye-wear}  & {$\times$}  & {$\times$}   &  \checkmark & {$\times$} &  \checkmark              \\ [1ex]
\hline
\textbf{Lacrimal caruncle} & {$\times$}      &     {$\times$}    &     {$\times$}        &    {$\times$}         &     \checkmark          \\ [1ex]
\hline
\textbf{Variable eyelid position}   &   {$\times$}      &{$\times$}       &     \checkmark        &    \checkmark         &     \checkmark          \\ [1ex]
\hline
\textbf{Real-time rendering}    &   \checkmark       &{$\times$}       &     {$\times$}       &    {$\times$}         &     {$\times$}          \\ [1ex]
%\hline
\end{tabular}
\caption{Comparison between existing synthetic rendering pipelines and datasets.}
\label{tbl:synthetic_datasets}
\end{table}

\section{Head Models}
Our rendering platform currently incorporates 24 head models with varying skin color, gender, and eye shape. Figure~\ref{fig:head_models} provides several example near-eye renderings. Head models were purchased from an online repository\footnote{\url{https://www.3dscanstore.com/}}. The associated textures contain 8K color maps captured using high-quality cameras. Eighteen models were selected to capture diversity (9 male, 9 female) in generating a training set, while the remaining 6 models (3 male and 3 female) were used to generate a testing set. To approximate the properties of human skin in the infrared domain, the red channel from the original diffuse texture map is incorporated for rendering purposes. 

%In order to match the visual properties of the skin under infrared imaging, we only used the red channel of the original diffused texture map of the models. 

To overcome the challenge of controlling the placement of eyelashes relative to the eye-lid position, we replaced each of the model's original eyelashes with Blender's built-in hair particle editor which provides a plausible physical simulation of hair behavior. This is similar to the approach used by Swirski et al.~\cite{Swirski2014RenderingEvaluation}. We also replaced the basic 3D eyeball included with the 3D head models with our own customized 3D eyeball that provides greater control and more faithfully simulates the structure of real eyes.

% To handle issues such as difficulty imaging/replicating hair andeyelashes patterns, the model’s original eyelashes were removedand replaced using Blender’s built-in hair particle editor in a mannerfaithful to the eyelid deformation caused by the eye gaze position

\section{Eye Model}
Reconstructing all parameters that influence the imaging of a person's eye is a difficult task, and it is common to make simplifying assumptions regarding its structure. We used a modified Emsley-reduced optical model of the human eye~\cite{Atchison2016OpticalEye,Dierkes2018ARefraction}. Table~\ref{tbl:eye_features} summarizes the various basic physical properties. Modeling and rendering were accomplished using Blender-2.8.
 
\begin{table}[h]
\begin{center}
\begin{tabular}{|c|c|c|}
\hline
%\textbf{Feature} & \textbf{Radius (mm)} & \begin{tabular}[c]{@{}l@{}}\textbf{Refractive}\\ \textbf{index (n)}\end{tabular}\\
\textbf{Feature} & \textbf{Radius (mm)} & \textbf{Refractive index (n)}\\
\hline
Cornea & 7.8 mm & 1.3375 \\
\hline
Pupil & 1-4 mm & $\times$  \\
\hline
Iris disc  & 6 mm & $\times$\\
\hline
Eyeball sphere & 12 mm & $\times$\\

\hline
\end{tabular}
\end{center}
\caption{Basic physical properties of our eye model. Radius and refractive index values courtesy of Dierkes et al.~\cite {Dierkes2018ARefraction}}
\label{tbl:eye_features}
\end{table}

\begin{figure} [!h]
\begin{center}
\includegraphics[width=1\columnwidth]{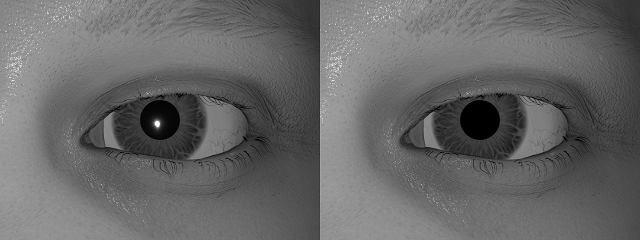}
\end{center}
   \caption{Comparison renderings to illustrate improvements offered by our model. With tear film (left). Without (right).}
\label{fig:corneal_reflection}
\end{figure}

\noindent Furthermore, our eye model incorporates the following features: 
\paragraph{\textbf{Tear film:}} Similar to previous work~\cite{Wood2016LearningImages}, we designed a tear film on the outermost surface of the eyeball with glossy and transparent properties to produce plausible environmental reflections on the surface of the eye (see Figure~\ref{fig:corneal_reflection}).

\begin{figure} [!h]
\begin{center}
\includegraphics[width=1\columnwidth]{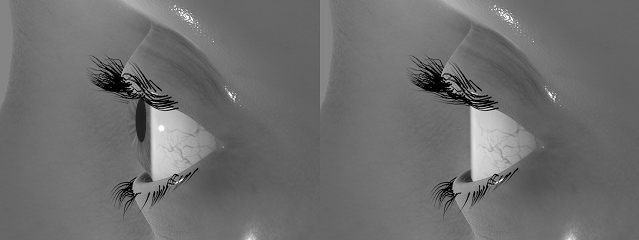}
\end{center}
   \caption{Comparison renderings to illustrate improvements offered by our model. With aspherical cornea (left). With no cornea (right).}
\label{fig:cornea}
\end{figure}

\paragraph{\textbf{Aspherical cornea:}} In contrast to previous work, we chose to render a physiologically accurate corneal bulge (see Figure~\ref{fig:cornea}). The corneal topography is modeled as a spheroid, $x^2+y^2+(1+Q)z^2-2Rz=0$~\cite{Durr2015CornealSubstitutes}, where $Q$ is the asphericity and $R$ is the corneal radius of curvature. Research has shown that the human eye exhibits $Q$ value of $\mu=-0.250$, $\sigma = 0.12$~\cite{Durr2015CornealSubstitutes}. Our corneal models incorporate three asphericity values, -0.130, -0.250 and -0.370, which were represented uniformly during rendering.

\begin{figure} [!h]
\begin{center}
\includegraphics[width=1\columnwidth]{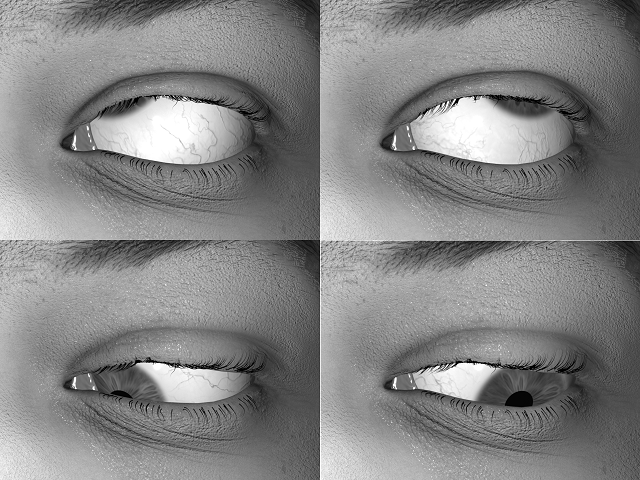}
\end{center}
   \caption{Renderings to illustrate improvements offered by our model. Eye lid deformation (shown at extreme gaze angles).}
\label{fig:eyelid_deformation}
\end{figure}

\paragraph{\textbf{Deformable eyelids:}} In order to avoid any visible gaps between the eyelids and the eyeball, the original 3D vertices in each eye socket were morphed to a snug fit around our custom eyeball. Using Blender's inbuilt wrapping function, we deformed the eyelid mesh to conform to the corneal contour below it. To mimic human behavior, the amount of eyelid closure was approximated by a linear function of eye rotation in the vertical axis (see Figure~\ref{fig:eyelid_deformation}).

% In order to have better control of the individual components of the eye like the iris, sclera etc we modeled own eye ball rather than the using the eye ball from the original scan. So in order to avoid any visible gaps between the eye lids and the eye ball, the original 3D mesh vertices were moved to perfectly fit the custom eye ball. Using the Blender's inbuilt wrapping function we deform the eye lid mesh to flow around the eye ball. The eye lid motion is a linear function of eye rotation in the vertical axis providing a realistic representation of an actual human eye region.

%\sout{This results in a natural faded appearance to the iris color.}

\begin{figure}[h]
\begin{center}
\includegraphics[width=1\columnwidth]{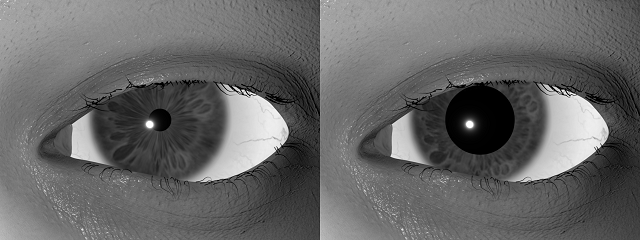}
\end{center}
   \caption{Renderings to illustrate improvements offered by our model. Variable size pupil aperture. 1 mm radius(left) to 3 mm radius(right).}
\label{fig:pupil}
\end{figure}

\paragraph{\textbf{Pupil aperture:}} Previous datasets have modeled the pupil as an opaque black disc. The pupil in our eye model was accurately modeled as an aperture such that constriction or dilation of the pupil was accompanied by appropriate deformation of surrounding iris texture (see Figure~\ref{fig:pupil}). The pupil aperture opening was uniformly distributed between 1 mm to 4 mm in radius~\cite{Dierkes2018ARefraction}.

\begin{figure}
\begin{center}
\includegraphics[width=1\columnwidth]{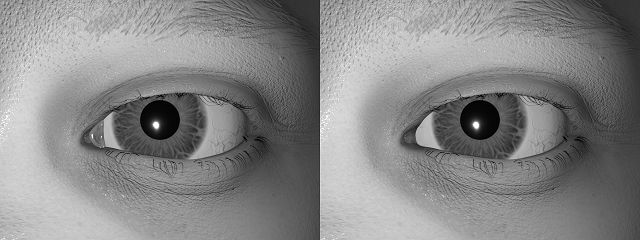}
\end{center}
   \caption{Comparison renderings to illustrate improvements offered by our model. With lacrimal caruncle (left). Without (right). }
\label{fig:carnuncle}
\end{figure}

\paragraph{\textbf{Lacrimal caruncle:}} In contrast to previous works, we included the lacrimal caruncle, a small pink nodule positioned at the inner corner of the eye (see Figure~\ref{fig:carnuncle}). The lacrimal caruncle consists of skin, sebaceous glands, sweat glands, and hair follicles. Hence when generating the segmentation mask of the respective synthetic eye image, the lacrimal caruncle was considered to be part of the skin (see Figure)~\ref{fig:eye_mask}) as opposed to OpenEDS~\cite{Garbin2019OpenEDS:Dataset}, which segmented lacrimal caruncle as a part of sclera. 

\begin{figure} [!h]
\begin{center}
\includegraphics[width=1\columnwidth]{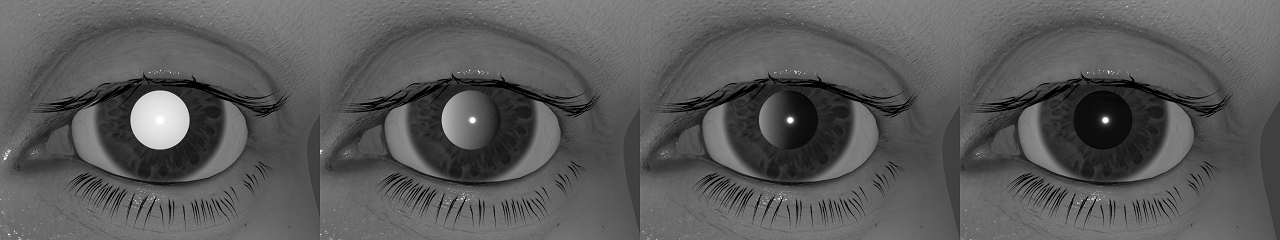}
\end{center}
   \caption{Renderings to illustrate improvements offered by our model. Bright pupil effect at varying degrees of angular separation between the imaging optical axis and the light source. From left to right: 0$^\circ$, 1.16$^\circ$, 1.51$^\circ$, 2.25$^\circ$.}
\label{fig:retroref}
\end{figure}

\paragraph{\textbf{Bright pupil response:}} The bright pupil response occurs when a light source is within $\sim$2.25$^\circ$ of separation from the imaging optical axis~\cite{Nguyen2002DifferencesEyes}. To simulate eye physiology, we added retroreflectivity to the retinal wall.Reflectivity increases as the angle of separation decreases following a Beckmann distribution (see Figure~\ref{fig:retroref}).

\begin{figure}
    \centering
    \includegraphics[width=1\columnwidth]{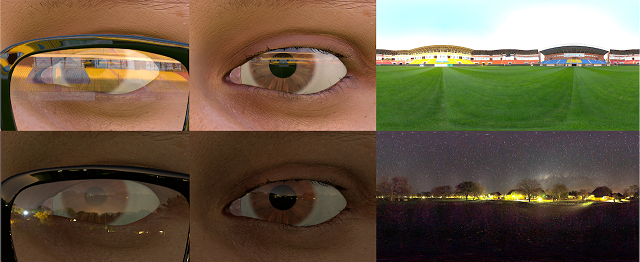}
    \caption{Renderings with (left) and without (middle) glasses. Corresponding HDR environment map (right).} 
    \label{fig:glasses_HDRenv}
\end{figure}

\paragraph{\textbf{Environment mapping and reflective eye-wear:}} Following Wood et al.~\cite{Wood2016LearningImages}, we used 360$^\circ$ HDR images to simulate the reflections from the external environment. The environment texture is mapped onto a sphere with the eye model at its center. Each pixel intensity on the texture acts as a separate light source that illuminates the model. We used 25 HDR images obtained from an online repository.\footnote{\url{https://hdrihaven.com/}} Of these, nine were indoor, and 16 were outdoor scenes (see Supplementary Material). The pixel intensity of the environment texture was varied between $\pm$50$\%$ of its original value. Textures were chosen at random, and their global intensity scaled to ensure equal proportions of dark, well lit, and saturated imagery. The environment map was rotated up to 360$^\circ$ along the y-axis and up to $\pm$60$^\circ$ on the x and z-axis to induce a unique reflection pattern on the model for every rendered image. Figure~\ref{fig:glasses_HDRenv} shows examples of environment mapping with and without reflective eye-wear.

% \noindent \textbf{Pupil size variation} In previous work, the pupil is typically modeled as a black disc, hence the influence of pupil constriction and dilation upon the visual appearance of the surrounding iris is purely due to occlusion. By contrast, in our work, the pupil is accurately modeled as an aperture in the iris such that constriction or dilation of the pupil is accompanied by appropriate deformation of the surrounding texture of the iris.\\

% \noindent The top row of Figure~\ref{fig:all_effects} provides example renderings with all of these features enabled. For comparison, the bottom row shows the renderings obtained when these features were disabled.
\paragraph{\textbf{Iris and sclera textures}}
\label{sec:rendering:subsec:texturecomb}
Our rendering platform currently incorporates 9 infrared (IR) textures of the iris 7 obtained using IR photography of the human eye (courtesy of John Daugman~\cite{Daugman2009HowWorks}) and 2 artificial renderings previously used by Swirksi et al.~\cite{Swirski2015GazeDisplays}. Note that among the 7 photographed textures of the iris, parts may be occluded due to eye lashes, eye lid position, or by reflections. In order to remove these artifacts, the images were manually edited using Photoshop. The texture for the sclera was purchased from an online repository~\footnote{\url{https://www.cgtrader.com/}}. Since, we only had access to one sclera texture and 9 iris textures, a random rotation between 0$^\circ$ and 360$^\circ$ was applied to the sclera and iris textures to increase variability in the rendered eye images.

\section{Rendered Datasets}
\label{sec:rendered_datasets}

\begin{figure}
\begin{center}
\includegraphics[width=1\columnwidth]{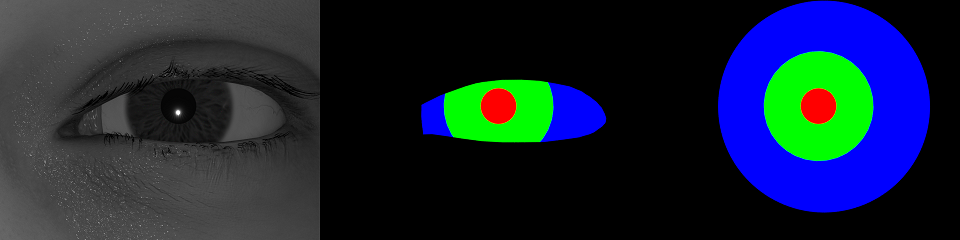}
\end{center}
   \caption{Sample synthetic image along with ground truth mask of pupil (red), iris (green), and sclera (blue) with and without skin. }
\label{fig:eye_mask}
\end{figure}

Three new datasets were rendered for use in the training of multiple independent (CNNs) for the semantic segmentation of eye features, and to test the ability for these CNNs to generalize across datasets collected using different configurations of the eye, camera, and infrared emitter(s). Two of the new datasets, \textit{S-NVGaze} (see Figure~\ref{fig:snvgaze_img}) and \textit{S-OpenEDS} (see Figure~\ref{fig:sopeneds_img}), are synthetic renderings intended to mimic the synthetic NVGaze~\cite{Kim2019NVGaze:Estimation} and real OpenEDS~\cite{Garbin2019OpenEDS:Dataset} datasets. The third dataset, \textit{S-General} (see Figure~\ref{fig:sgeneral_img}) reflects a wide distribution of possible camera positions and orientations in the Pupil Labs Core mobile eye tracker. Each new dataset includes 51,600 path-traced images rendered using Blender's Cycles rendering engine for a total of 154,800 images. 
%For the purpose of testing and validation against of S-NVGaze, S-OpenEDS, and S-General against the original NVGaze and OpenEDS datasets, all images were rendered only in the infrared domain.\\

Eivazi et al.~\cite{Eivazi2019ImprovingAugmentation} augmented a dataset of real eye images by recording scene reflections off the anterior side of black coated glasses and superimposing them on the real eye imagery. To achieve similar reflections, we incorporate a 3 mm thick eyeglasses with black frames. Half the images in each dataset were rendered with eyeglasses.

Eye pose was uniformly distributed within $\pm$ 30$^\circ$ in both azimuth and elevation.  For each rendered image in the dataset, we generated ground truth masks of the sclera, iris, and pupil with and without the skin (see Figure~\ref{fig:eye_mask}) to facilitate evaluation of new and existing eye-tracking algorithms. We record additional metadata, including the 2D and 3D center of various eye features relative to the camera, as well as eye pose in degrees and the eye camera intrinsic matrix. A comprehensive list of the properties of the various datasets is provided in Table~\ref{tbl:dataset_properties}.

\begin{table*}[t]
\begin{center}
\begin{tabular}{|c|c|c|c|c|c|}
\hline
\textbf{Dataset} & \textbf{Number Images} & \textbf{Resolution} & \textbf{Number of Subjects}& \textbf{Camera distance} & \textbf{Number of emitters} \\
\hline
NVGaze  & 2M &  1280 x 960 & 10 (5) &  $\times$ &    4  \\
\hline
S-NVGaze (ours) & 51,600 & 640 x 480 & 24 (12) & 3.5 cm to 4.5 cm &    1   \\
\hline
 OpenEDS  & 12,759  & 400 x 640 & 152 (82) & $\times$ &   16*   \\
\hline
 S-OpenEDS (ours)  & 51,600 & 400 x 640 & 24 (12) & 3.5 cm to 4.5 cm &   16  \\
 \hline
  S-General (ours)  & 51,600 & 640 x 480 & 24 (12) & 2.5 cm to 4.5 cm &   1 \\
\hline
\end{tabular}
\end{center}
\caption{Comparison of the image properties and camera setting used in NVGaze, S-NVGaze, OpenEDS, S-OpenEDS, S-General. The $\times$ symbol denotes a property that was either not reported, or not applicable. () indicated the number of female subjects. *Count based on number of corneal reflections.}
\label{tbl:dataset_properties}
\end{table*}

\subsection{S-OpenEDS}
\begin{figure}[!h]
\begin{center}
\includegraphics[width=1\columnwidth]{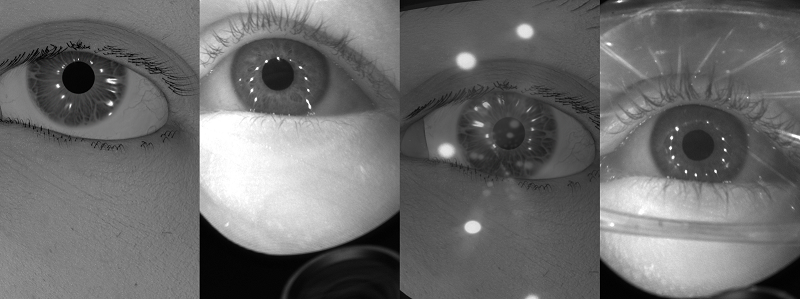}
\end{center}
   \caption{Comparison of images from S-OpenEDS (odd columns) with corresponding images from OpenEDS (even columns).}
\label{fig:sopeneds_img}
\end{figure}

OpenEDS is a dataset containing 12,759 off-axis images of real eyes that were captured with IR eye cameras positioned within a head mounted display~\cite{Garbin2019OpenEDS:Dataset}. To mimic the appearance of this dataset, we approximate eye camera orientation by overlaying and comparing segmentation masks. In order to approximate the lighting conditions of the OpenEDS dataset, 16 point-sources were arranged near the virtual eye camera in a circular pattern. Blender's native compositor was used to imitate the resulting reflection pattern. The camera position is uniformly shifted $\pm$5 mm in the horizontal axis and vertical axis (see Figure~\ref{fig:cam_pose}) to simulate slippage of the HMD. The camera was rotated -10$^\circ$ along the elevation and the distance was uniformly varied from 3.5 cm to 4.5 cm from the tip of the eye based on empirical observations. Figure~\ref{fig:sopeneds_img} shows side-by-side comparisons of images from OpenEDS and S-OpenEDS.

% We have focused on the 12,759 real eye images along with their segmented masks. 
%16 point-source light sources were arranged around the virtual eye camera in circular pattern. The radius of the circle was set to 9 cms,
%$\pm$5 mm in the azimuthal and elevation planes in-order to account for the slippage of head-mounted camera. The camera distance was uniformly varied from 3.5 cm to 4.5 cm. 

% To replicate the circular reflections on the eye from the OpenEDS dataset, 16 \gd{point-source illuminants} \sout{light sources} were positioned around the camera while rendering. Blender's inbuilt compositor settings were used to imitate the starburst reflection patterns upon the iris from the illuminants. The camera position was uniformly shifted $\pm$5 mm in the horizontal and the elevation plane in-order to account for the slippage of head-mounted camera. The camera distance was uniformly varied from 2.5 cm to 4.5 cm.  

\subsection{S-NVGaze}
\begin{figure}
\begin{center}
\includegraphics[width=1\columnwidth]{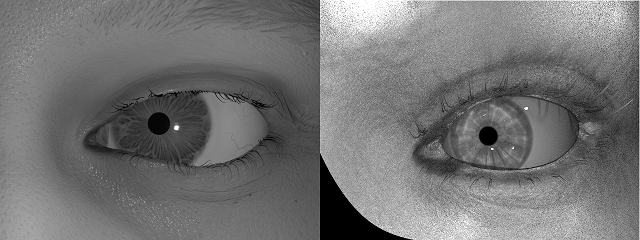}
\end{center}
   \caption{Comparison of image from S-NVGaze (left) with corresponding image from NVGaze (right).}
\label{fig:snvgaze_img}
\end{figure}

The NVGaze dataset contains 2M synthetic eye images generated using an on-axis camera configuration. For S-NVGaze, we rendered images at a resolution of 640 x 480 pixels with 200 rays per pixel. Similar to S-OpenEDS, we placed the camera in front of the eyes within an empirically derived distance of 3.5 cm to 4.5 cm from the tip of the eye. Eye camera position was varied $\pm$5 mm along the vertical and horizontal axes (see Figure~\ref{fig:cam_pose}) to match the simulated headgear slippage conditions observed in the NVGaze dataset. One point light source was placed near to the camera. Figure~\ref{fig:snvgaze_img} shows a side-by-side comparison of an image from NVGaze with the corresponding image from S-NVGaze.

% were placed 2 mm to the anterior and nasal side of the eye camera.

%A single point-source light source was used 2 mm to the left and 2 mm below the camera position instead of 16 as in S-OpenEDS. Note that the difference of S-NVGaze with S-OpenEDS are the number of light source single point-source vs 16 point source and their positions , image resolution of 640px x 480px vs 400px x 640px,  no camera rotation vs 10$^\circ$ camera rotation along x-axis respectively.

%2 mm to the left and 2 mm below the camera position.} in this dataset instead of 16 as in S-OpenEDS.

% The Nvgaze dataset contains 2M images of synthetically generated eye images. On-axis camera configurations were used to generate the images which provided better image quality. The images were rendered at a resolution of 640 x 480 pixels with 200 rays per pixel. The camera was place in front of the eyes with a varying distance of 2.5cm to 4.5cm . Similar to S-OpenEDS the camera position was varied $\pm$5 mm in vertical and horizontal axis. Only a single light source in this dataset instead of 16 as in S-OpenEDS.

\subsection{S-General}
\begin{figure}[!h]
\begin{center}
\includegraphics[width=1\columnwidth]{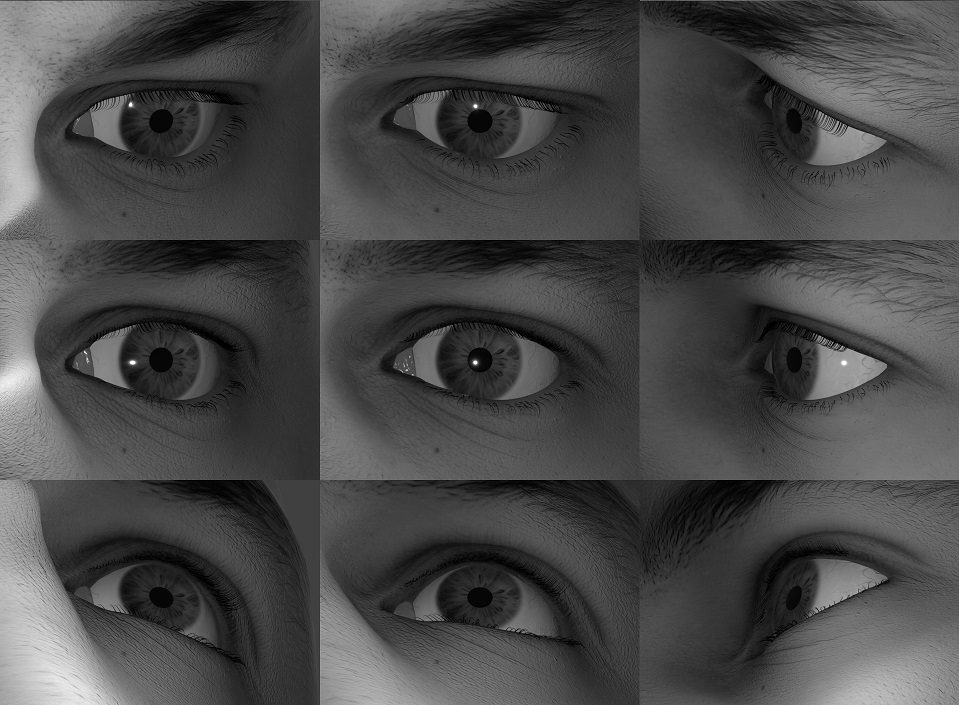}
\end{center}
   \caption{Sample images form S-General showing eye with fixed gaze and varying camera position. The camera position at -20$^\circ$, 0$^\circ$, 60$^\circ$ (left to right) in azimuthal plane and -20$^\circ$, 0$^\circ$, 40$^\circ$ in elevation plane (top to bottom).}
\label{fig:sgeneral_img}
\end{figure}

\begin{figure}
\begin{center}
\includegraphics[width=1\columnwidth]{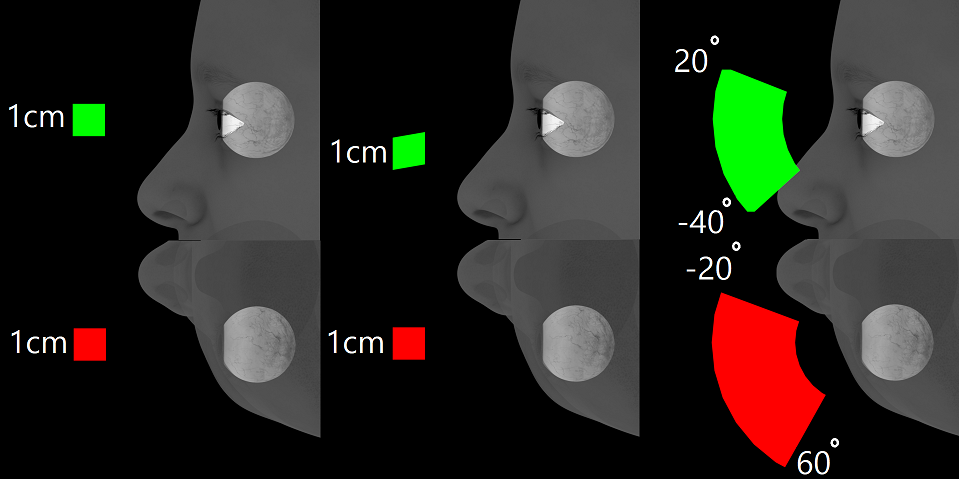}
\end{center}
   \caption{Camera positions used for S-NVGaze (left column), S-OpenEDS (middle column), and S-General (right column). Side-view (top). Top-view (bottom).}
\label{fig:cam_pose}
\end{figure}

This dataset approximates the conditions imposed by the Pupil Labs Core mobile eye tracker. The camera position was uniformly distributed within an eye-centered spherical manifold subtending -20$^\circ$ to 60$^\circ$ along the azimuth and -20$^\circ$ to 40 $^\circ$ in elevation (see Figure~\ref{fig:cam_pose}), which encompasses the range of camera positions afforded by the Pupil Labs Core mobile eye tracker. A smaller jitter of $\pm1$ mm in the vertical and horizontal plane is added to account for possible variation due to eye tracker slippage. Figure~\ref{fig:sgeneral_img} shows example images of a fixed gaze with varying camera positions from the S-General dataset.

% The major difference to S-NVGaze and S-OpenEDS is a wider variation in camera position and orientation. Additionally, only one point-source light source is used as in original S-NVGaze to replicate most of the available commercial eye-trackers.

%Note that the difference with S-NVGaze and S-OpenEDS is the camera distance was uniformly varied from 2.5 cm to 4.5 cm and its orientation was varied from -20$^\circ$ to 60$^\circ$ azimuthal and -20$^\circ$ to 40 $^\circ$ in elevation. Only a single light source camera was used in S-General whereas in S-OpenEDS 16 were used.

% In contrast with S-NVGaze and S-OpenEDS, the camera center was not co-linear with \sout{the pupil center}, but the camera axis was shifted 1 mm in vertical and horizontal plane in-order to account for the camera miscalibration. Similar to S-NVGaze only one light spurce was used.

% The camera was positioned within a range of 3 cm to 15 cm from the eyeball center, and oriented so that its optical axis lies on the cornea center.

% \gd{What is the position fo the camera?} Orientation of the eye camera is uniformly sampled within a range of 45 degrees in azimuth and 75 degrees in elevation around the . We randomly position the eye camera 

% Blender's inbuilt Cycles render engine was used to render images at a 640 by 480 pixel resolution with 500 rays per pixel. In this work, we share two groups of rendered images, RIT-Eyes-IR and RIT-Eyes-RGB. Each group contains three different head models as listed previously with 3980 images in each. This culminates in 3980 (images/head model) $\times$ 3 (heads models/group) $\times$ 2 (groups) $=$ 11940 images. For every image in the dataset, we record two masks which assigns each pixel to a relevant category (sclera, iris or pupil). The masks are generated with and without the presence of skin. Parameters such as 3D eye ball, iris, pupil and sclera center are recorded along with the eye pose and camera intrinsic matrix. The eye camera spans a range of 45 degrees in azimuthal and 75 degrees in elevation plane in front of the eye. We randomly position the eye camera within a range of 3 cm to 15 cm from the eye ball center and orient the eye camera such that it's optical axis lies on the cornea center.

\section{Model Architecture}
\label{sec:model_arch}

Two architectures were used when testing the ability for models to generalize  across datasets collected using differing configurations of the eye, camera, and emitter(s):
RITnet~\cite{Chaudhary2019RITnet:Tracking} (0.25M parameters) and SegNet~\cite{Badrinarayanan2017SegNet:Segmentation} (24.9M  parameters).

%RITnet has shown state-of-the-art results on semantic segmentation of eye images for OpenEDS dataset in memory-constrained settings. Garbin et al. have shown SegNet architecture to produce a baseline result for the same OpenEDs dataset without memory-constrained settings.

\paragraph{\textbf{Dataset:}} Each dataset (S-OpenEDS, S-NVGaze, and S-General) consists of 39600 training images (36000 open eye cases, 1800 random eyelid position cases ranging from 80\% to $<$100\% closure and 1800 completely closed eye cases) of 18 head models and 12000 testing images of 6 different head models. Each training set is further divided into a 80-20 training/validation stratified split based on binned pupil center locations of each image.

To derive conclusions about varying eye camera poses and domain generalizability, each of the previously mentioned neural network architectures (RITnet and SegNet), are trained under three different configurations: a) training with S-NVGaze, b) training with S-OpenEDS, and c) training with a combination of randomly selected S-NVGaze and S-OpenEDS images (50$\%$ each). A subset of images was set aside for model testing, including the entirety of the S-General dataset, 1500 random images extracted from each head model of the NVGaze dataset (60k images), and 2392 images present in the official OpenEDS validation set.

%As mentioned in earlier section, our training paradigm is divided into three dataset combinations.  Each 

%dataset consists of 2000 images of 18 individuals with varying head model properties.  We test our model and the dataset generalizability in the two publicly available dataset NVGaze (40000/) and OpenEDS (validation set). Note that OpenEDS test set labels are not publicly available yet. Furthermore, we report the accuracy's in the six additional synthetic head models dataset for cases consisting of S-NVGaze, S-OpenEDS, and general synthetic images at random camera position and gaze directions.

\paragraph{\textbf{Training:}} We trained our models using the Adam optimizer~\cite{Kingma2014Adam:Optimization} with a learning rate of 5e$^{-4}$ for 40 epochs. We reduced the learning rate by a factor of 10 when the validation loss plateaued for more than 5 epochs. Both models are trained using the loss function strategy proposed in RITnet~\cite{Chaudhary2019RITnet:Tracking}. This strategy involves using a weighted combination of four loss functions:

\begin{itemize}
    \item cross-entropy loss $\mathcal{L}_{CEL}$
    \item generalized dice loss~\cite{Sudre2017GeneralisedSegmentations} $\mathcal{L}_{GDL}$
    \item boundary aware loss $\mathcal{L}_{BAL}$
    \item surface loss~\cite{Kervadec2018BoundarySegmentation} $\mathcal{L}_{SL}$
\end{itemize}

The total loss $\mathcal{L}$ is given by a weighted combination of these losses as $\mathcal{L} = \mathcal{L}_{CEL} (\lambda_1 + \lambda_2 \mathcal{L}_{BAL} ) + \lambda_3 \mathcal{L}_{GDL} + \lambda_4 \mathcal{L}_{SL}$. In our experiments, we used $\lambda_1 = 1$, $\lambda_2 = 20$, $\lambda_3 = $ (1 – $\alpha$) and $\lambda_4 = $ $\alpha$, where $\alpha = epoch/125$.

\paragraph{\textbf{Image Augmentation:}}
Image augmentation aids in broadening the statistical distribution of information content and combats overfitting~\cite{Eivazi2019ImprovingAugmentation}. Previous efforts~\cite{Chaudhary2019RITnet:Tracking, Park2018LearningSettings} have shown that data augmentation on eye images improves the performance of convolutional networks under naturalistic conditions such as varying contrast, eye makeup, eyeglasses, multiple reflections and image distortions. In this work, we utilize the following image augmentation schemes:

\begin{enumerate}
    \item Since our work contains left-eye images exclusively, images were flipped about the vertical axis to simulate right eye conditions
    \item Image blurring using a Gaussian kernel (width = 7 pixels, $\sigma$ = 2 - 7 pixels)
    \item Thin lines drawn around a random center (120 < x < 280, 192 < y < 448)~\cite{Chaudhary2019RITnet:Tracking} to simulate glare on glasses
    \item Gamma correction with one of the following random factors: 0.6, 0.8, 1.2, or 1.4
    \item Intensity offset up to $\pm$25 levels)
    \item Down-sampling followed by the addition of random Gaussian noise (mean=0, $\sigma$ = 2 to 16 levels) and up-sampling by factor of 2-5 
    \item No augmentation
\end{enumerate}

Each augmentation scheme occurs with 14$\%$ probability.

% TODO: add error bars to the figure
\begin{figure}[ht]
\begin{center}
\includegraphics[width=1\columnwidth]{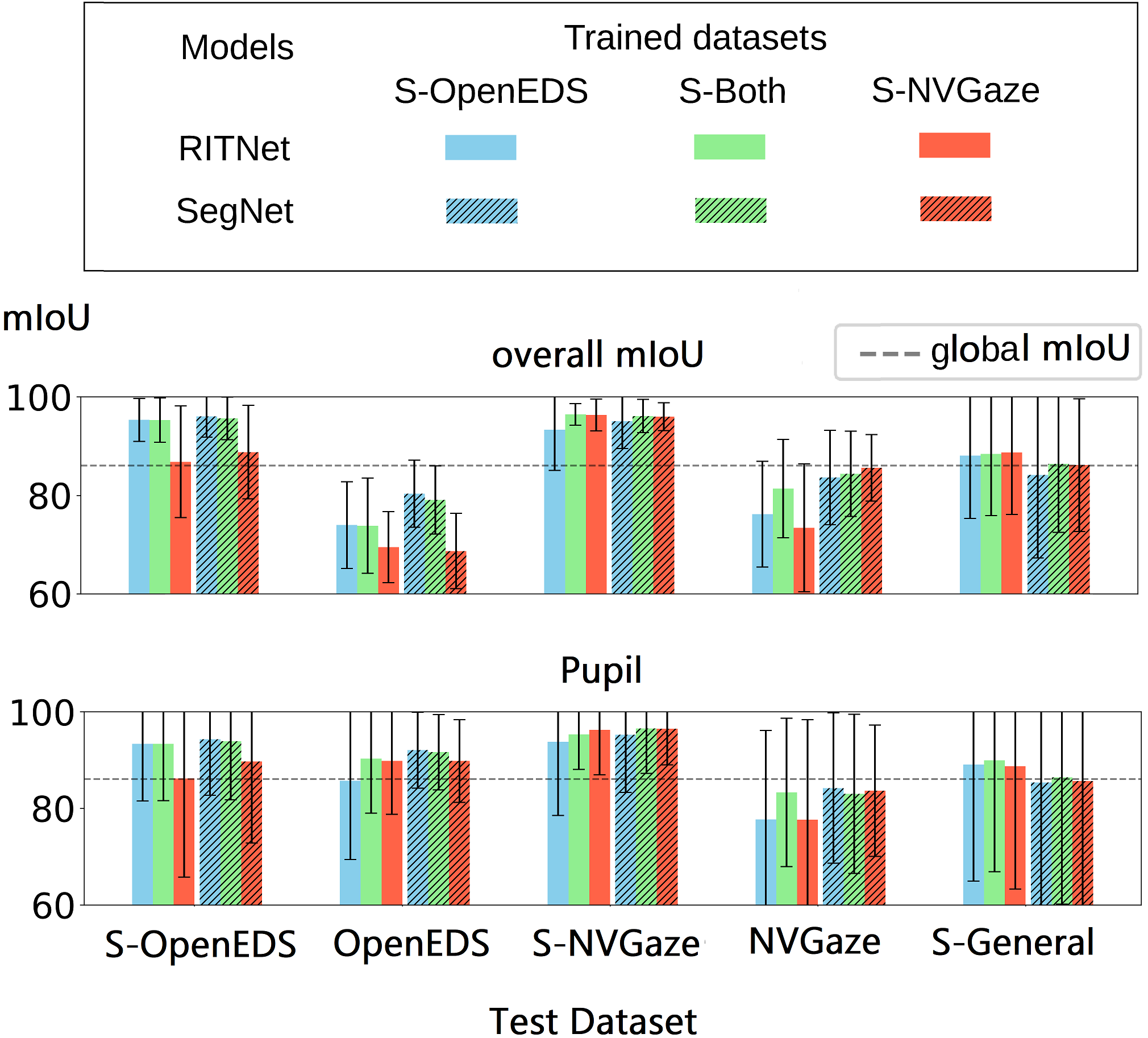}
\end{center}
 \caption{Results for overall mIoU and pupil-class mIoU. The horizontal axis consists of number of vertical blocks representing various test-dataset (S-OpenEDS, OpenEDS, S-NVGaze, NVGaze and S-General). Each block represents the performance of the particular dataset under various training conditions and models. Each block is further divided into unshaded and shaded blocks representing two model architectures, RITnet and SegNet respectively. Each of the unshaded/shaded block is further divided into three columns (S-OpenEDS (left), S-Both (middle) and S-NVGaze (right). Each row indicate different cases (the overall mIoU and pupil class mIoU). The error bars show the standard deviation of the respective scores.}
\label{fig:iou_plot}
\end{figure}

\section{Results and Discussion}
\label{sec:result_discussion}

Performance is evaluated using the \textit{mean Intersection over Union (mIoU)} metric (in \%) for both models in different test conditions for each class (pupil, iris, sclera, and background) and for all classes combined (which we refer to as the \textit{overall mIoU}).

Results for \textit{overall mIoU} and the pupil-class are summarized in Figure~\ref{fig:iou_plot}. Overall, the models performed quite well, with a \textit{global mIoU} score (mean of all mIoU scores, indicated by the dotted line) of \textbf{86.06}. However, there is a large standard deviation in the overall mIoU (8.52) with scores ranging from 77.54- 94.59. For the pupil class, the standard deviation was 5.17
with a range of (84.09-94.43). Results for the other semantic classes (iris, sclera, and background) are included in the Supplementary Materials. 

Below, we test several hypotheses related to model performance through targeted comparisons of training/testing data. For the sake of reference throughout the discussion section, and to better understand fluctuations in model performance, performance on these specific tests will be compared to the global mIoU.\\

% Global mIoU is the mean of every overall mIoU scores which is used for comparison of our test scores.  

% The figure also shows the performance of the pupil....

% Accuracy for the other semantic classes (background, sclera, and iris) are reported in Supplementary materials.

% Do CNN trained on our datasets perform well on when testing on reserved test data from the same dataset?
% Peak model performance is approximated when a CNN trained with synthetic eye imagery is applied to synthetic data drawn from the same distribution of rendering properties.

\noindent \textit {\textbf{Do the CNNs generalize to head models not present during training?}} Tests reveal that models performed well above global mIoU when the CNNs were trained and tested using different head models, while keeping the manifold of camera positions/orientations constant across training and testing. This suggests that a CNN specifically trained for use with a particular eye tracker can generalize well to previously unseen faces that differ in both structure and appearance from the training set. Both S-OpenEDS and S-NVGaze achieved mIoU scores above 95 when tested on head models not used during training, as shown in Table~\ref{tbl:iou_best} (highlighted in red).\\

\noindent \textit{\textbf{How does the range of camera poses represented by the training dataset affect the ability to generalize to new camera poses?}} Our results demonstrate that the CNNs can generalize beyond the spatial area in which they were trained.  This conclusion is supported by the fact that when the CNNs were tested on datasets in which the spatial distribution of camera positions/orientations were distinct from that represented by the training dataset, we observed performance at levels equal to or above the global mIoU. As shown in Table~\ref{tbl:iou_best} (highlighted in green), when a model trained on S-OpenEDS was tested on S-NVGaze, the scores were 93.32 with RITnet and 95.04 with SegNet, and when trained on S-NVGaze and  tested on S-OpenEDS, the scores were 86.81 with RITnet, 88.75 with SegNet.

Our models also demonstrated an ability to  generalize across two distinct distributions of camera poses represented in the training set without a need to increase its size. There was no appreciable degradation of mIoU score when training using a dataset of the same size, with half of the images coming from S-OpenEDS and half coming from S-NVGaze, and testing with either S-OpenEDS or S-NVGaze (see Table~\ref{tbl:iou_best}, highlighted in red). Since there was no drop in performance relative to training and testing the datasets on subsets of their own imagery, we can conclude that the 39,600 image training dataset was a sufficiently dense sampling to account for simulated eye imagery taken from a camera position falling within the manifold spanning the regions represented in the first two columns presented in Figure~\ref{fig:cam_pose}.\\

\begin{table}[t]
\begin{tabular}{|p{2cm}|p{1.7cm}|p{1.7cm}|p{1.7cm}|}
\hline

%% previous table format with best performing model and score difference %%
% \textbf{TEST \textbackslash{} TRAIN } & \textbf{S-NVGaze  }       & \textbf{S-OpenEDS  }       & \textbf{S-NVGaze $+$  S-OpenEDS }  \\ \hline
% S-NVGaze                  & \textbf{R}/96.86/0.24 & \textbf{S}/96.02/1.79 & \textbf{S}/97.01/0.01                         \\  \hline
% NVGaze                    & \textbf{S}/88.24/7.21 & \textbf{S}/86.03/5.98 & \textbf{S}/86.65/4.15 \\  \hline
% S-OpenEDS                 & \textbf{S}/90.16/2.3  & \textbf{S}/96.71/0.73 & \textbf{S}/96.44/0.45                         \\  \hline
% OpenEDS                   & \textbf{R}/69.69/1.0  & \textbf{S}/80.6/7.34  & \textbf{S}/79.56/5.98 \\  \hline
% S-General                 & \textbf{R}/91.22/2.25 & \textbf{R}/90.58/3.33 & \textbf{R}/90.89/1.82 \\ \hline
% \end{tabular}
% \caption{The table represents the best performing model (\textbf{R}:RITNet/ \textbf{S}:SegNet) / best score/ score difference between the score of the two models on overall mIoU score when trained on S-NVGaze, S-OpenEDS and both (S-NVGaze + S-OpenEDS) and tested on S-NVGaze, NVGaze, S-OpenEDS, OpenEDS, S-General.}
% \label{tbl:iou_best}
% \end{table}

\textbf{TEST \textbackslash{} TRAIN } & \textbf{S-NVGaze  }       & \textbf{S-OpenEDS  }       & \textbf{S-NVGaze $+$ S-OpenEDS }  \\ \hline
S-NVGaze                  & \cellcolor{red} \textbf{96.31}/95.96 & \cellcolor{green} 93.32/\textbf{95.04} &\cellcolor{red}\textbf{96.41}/96.08                         \\  \hline
S-OpenEDS                 &  \cellcolor{green} 86.81/\textbf{88.75}  & \cellcolor{red}95.30/\textbf{95.99} & \cellcolor{red} 95.28/\textbf{95.62}                         \\  \hline
S-General                 & \cellcolor{blue}\textbf{88.71}/86.11 &\cellcolor{blue} \textbf{88.06}/84.11 & \cellcolor{blue} \textbf{88.40}/86.37 \\ \hline
NVGaze                    &  73.38/\textbf{85.57} & 76.17/\textbf{83.61} & 81.36/\textbf{84.36} \\  \hline
OpenEDS                   & \textbf{69.46}/68.67  & 73.94/\textbf{80.31}  & 73.81/\textbf{79.07} \\  \hline

\end{tabular}
\caption{Overall mIoU scores of the two models (RITnet / SegNet) when trained (listed on top row) on S-NVGaze, S-OpenEDS and both (S-NVGaze + S-OpenEDS) and tested (listed on the left column)  on S-NVGaze, NVGaze, S-OpenEDS, OpenEDS, S-General. The best performing model is indicated using bold font. The highlighted colors are used as an easy references to explain the hypotheses discussed in Section~\ref{sec:result_discussion}. (This figure is best viewed in color.)} 
\label{tbl:iou_best}
\end{table}

\noindent \textit{\textbf{Do the CNNs trained on one narrow spatial area generalize to a broader spatial area?}} Models are also able to perform at levels equal to or greater than the global mIoU when converged CNNs were tested against a synthetic dataset that represents a distribution of camera positions/orientations far larger than those present during training. When S-NVGaze (Figure~\ref{fig:cam_pose}, first column) was tested against S-General (Figure~\ref{fig:cam_pose}, third column), model performance reached an mIoU of 88.71 with RITnet, and 86.11 with SegNet. Similarly, when S-OpenEDS (Figure~\ref{fig:cam_pose}, middle column) was tested against S-General, model performance reached an mIoU of 88.06 with RITnet, and 84.11 with SegNet (see Table~\ref{tbl:iou_best} highlighted in blue).\\

%\sout{\noindent \textit{Do CNN trained on our dataset generalize to the dataset it was intended to mimic?}\\}
% We find that converged models result in scores lower or similar to the global mIoU when trained with synthetic approximations of previously published dataset and tested against the datasets they were intended to approximate. 

\noindent \textit{\textbf{Do the CNNs trained on our synthetic imagery generalize to the datasets they were intended to mimic?}} We examine how the mIoU scores change when testing on our own synthetic data vs. when testing on the original datasets we are attempting to mimic. When trained on S-NVGaze and tested on S-NVGaze and NVGaze, we observe that the mIoU scores dropped from (S-NVGaze) 96.31$\rightarrow$(NVGaze) 73.38 with RITnet and from  and (S-NVGaze) 95.96$\rightarrow$(NVGaze) 85.57 for SegNet. Similarly, when trained on S-OpenEDS and tested on S-OpenEDS and OpenEDS, we observe that the mIoU scores dropped from (S-OpenEDS) 95.30$\rightarrow$(OpenEDS) 73.94 for RITnet and (S-OpenEDS) 95.99$\rightarrow$(OpenEDS) 80.31 for SegNet (see Table~\ref{tbl:iou_best}).

One might speculate that this drop in performance is due to a poor match between the manifold of camera positions represented in the training dataset and the testing dataset. However all the trained CNN models performed better when tested on S-General (see Table 4 highlighted in blue), which has a much different range of camera positions and poses compared to the training datasets, than when the models are tested on the datasets they are trying to mimic. This suggests that the drop in performance may not be due to a mismatch in camera positions/orientations, but to differences related to appearance and pixel level statistics. A likely reason for this could be poor identification of scleral regions (mIoU of 34.42 for RITnet, and 49.04 for SegNet when training and testing on S-OpenEDS and OpenEDS respectively and mIoU of 47.74 for RITNet and 75.60 for SegNet when training and testing on S-NVGaze and NVGaze respectively). Since we only have one scleral texture, the failure to generalize may be due to a lack of variability in appearance. The fact that we also consider the lacrimal caruncle as belonging to the background while OpenEDS annotations consider it an extension to scleral regions may also be a contributing factor.\\

\noindent \textit{\textbf{Which model demonstrates better generalization?}} SegNet notably out performs RITnet when tested on OpenEDS ((RITnet) 73.94$\rightarrow$ (SegNet) 80.31) and NVGaze ((RITnet) 73.38$\rightarrow$ (SegNet) 85.57) after training on their synthetic counterparts. One cannot attribute these differences to loss functions, which were identical for all models presented here (note that these loss functions are different than those presented in~\cite{Badrinarayanan2017SegNet:Segmentation}). The difference in performance may be due to the fact that RITNet has significantly fewer parameters (0.25M) compared to SegNet (24.9M). \\

% Parameter count. (main reason)
% RITNet suitable for real time use up 300 Hz.
% TODO: check the value on the left side of all \rightarrow
% A possible reason is that the loss functions used with SegNet in this work (same ones as with RITnet but differ from the loss functions with the original SegNet paper) may aid SegNet in better capturing statistical variability. Another possible reason is the significantly large parameter count present in SegNet.

% \sout{as opposed to RITnet, a light weight and real-time segmentation architecture.}
% RITnet when tested on OpenEDS (73.26$\rightarrow$80.60) and NVGaze (81.03$\rightarrow$88.24) while trained on their synthetic counterparts. Previous work by Chaudhary\etal demonstrated that RITnet outperformed mSegNet (a computationally lighter variant to SegNet) by a significant margin in the OpenEDS challenge ~\footnote{\url{https://research.fb.com/programs/openeds-challenge/}}. This however is not reflected during our evaluations.

%\noindent \textit{Does pupil segmentation performance generalize well on the synthetic datasets?}
%\noindent \textit{\textbf{Pupil segmentation performance on synthetic datasets}}\\
\noindent \textit{\textbf{Do the CNNs generalize across our synthetic datasets when considering only the pupil region?}} The pupil is comparatively easier to segment than other eye regions and also it is the primary feature that is used to compute gaze direction in many current eye-trackers. We observe that pupil segmentation performance is indeed higher than the other eye regions (global mIoU for pupil/iris/sclera are 89.26/84.67/72.76 respectively). Furthermore, we observe that architectures trained on S-OpenEDS segment pupillary locations with higher accuracy on the S-NVGaze dataset (RITnet 93.76/SegNet 95.24) whereas the reverse (trained on S-NVGaze and tested on S-OpenEDS) is slightly degraded (RITnet 86.14 / SegNet 89.67). A potential reason for this behavior could be the difference in the number of glints present in S-OpenEDS (16) compared to S-NVGaze (1). Intuitively, if a network can accurately segment regions despite the presence of multiple glints, then that network would also perform well with fewer glints.

\section{Limitations and Future Work}
The models tested here suffered moderate drops in performance when trained on our synthetic imagery, and then tested on the original datasets, which the synthetic imagery was intended to approximate. There are several ways in which one might improve the ability to generalize to the original datasets. We might see performance gains by more accurately or fully accounting for a wider range of complicating visual features. For example, although our datasets include near-eye images with eyeglasses, the simulated glasses currently only reflect incoming light, they did not refract light.  Similarly, the presence of makeup, such as eye liner, eye shadow or mascara, which have been shown to interfere with many conventional algorithms for pupil detection and gaze estimation~\cite{Garbin2019OpenEDS:Dataset,Kim2019NVGaze:Estimation}, is not accounted for. Earlier works also do not address this.

% that, our current pipeline does not account for these artistic beauty effects like earlier works~\cite{Kim2019NVGaze:Estimation}.

% The dataset like earlier works~\cite{Kim2019NVGaze:Estimation}, did not systematically simulate the . %\nn{these datasets have subjects with eyeliner and mascara(real not synthetic) but they have not shown its effects}.

Although the experimental design here was sufficient to test several hypotheses related to the utility of synthetic imagery, there are also several ways in which our methodology can be improved upon. Our test was limited to few segmentation models and fixed hyper-parameters. A detailed exploration of multiple segmentation models and hyper-parameter settings might improve the results. We use a traditional train/test paradigm to evaluate architecture performance, and this paradigm is particularly sensitive to image selection. We attempted to alleviate this limitation by opting for a stratified sampling approach based on binned pupil center locations. This approach might be improved upon through the use of techniques such as double cross-validation. The sensitivity of tests might also be improved through the use of other metrics such as deviation in pupil centers.

The tests also suggest that the gap between synthetic and real image distributions could be one of the reasons why the CNN models trained on synthetic images could not generalize well to the real world imagery. Recently, advances in generative adversarial networks (GANs) have shown great promise for improving style transfer from one image to another. GANs have been used to refine the appearance of images from the Unity Eyes synthetic dataset~\cite{Shrivastava2017LearningTraining}. The improved appearance resulted in a smaller gaze error when compared to the unrefined images. Alternatively, it has been observed that the ability to generalize to real world imagery improves when a small number of hand-labelled real world eye images is included into the training set of synthetic eye imagery~\cite{Kim2019NVGaze:Estimation}. We plan to extend our rendering pipeline to leverage similar approaches.

%error fell  where  the  

%Theirs tests show that the eye gaze estimation performance of the CNN models when trained on the refined images improved by 21 as compared to the state-of-the-art on the MPIIGaze dataset\cite{MPGaze}.

%We plan to explore this approach in order to refine and improve the appearance of our synthetic imagery~\cite{Shrivastava2017LearningTraining}.

%We will attempt to improve generalizability through the use of style transfer using generative adversarial networks (GANs)~\cite{Shrivastava2017LearningTraining}. In \cite{Shrivastava2017LearningTraining}

%\sout{Limitations exist in our experimental} design. 

In contrast to previous synthetic rendering pipelines that generate temporally non-contiguous frames, our framework can generate sequences of eye movements similar to those in real-world datasets like Gaze-in-Wild~\cite{Kothari2020Gaze-in-wild:Activities} and 360em~\cite{Agtzidis2019AVideos}. We plan to explore if temporally contiguous eye images can be leveraged to improve the accuracy of gaze estimation algorithms.

%  including an accurate aspherical corneal model, a deformable iris, the lacrimal caruncle, gaze-coordinated blinks, and a retina with retroreflective properties, 

Finally, although our work provides improvements to various eye features, it is still based on a simplified eye model and thus there is room for improvement. Recent work has attempted to extract accurate 3D information about various features of the eye, including the sclera and iris, from high resolution imagery~\cite{berard2016lightweight, berard2014high, berard2019practical}. We plan to explore if incorporating such information in our rendering pipeline can enhance the visual appearance of the synthetic eye imagery and improve performance of gaze estimation.

%. Our syntheticeye image generation platform introduces several improvementsincluding an accurate aspherical corneal model, a deformable iris,the lacrimal caruncle (the small pink nodule located at the innercorner of the eye), gaze-coordinated blinks, and a retina with retro-reflective properties which can aid in developing ‘bright pupil’solutions [22] for head-mounted or remote eye trackers (see section3 and Table 1 for a comprehensive list of improvements

%
%Temporal

%, . We plan to explore if this temporal information can be leveraged to improve the accuracy of gaze estimation. 

%Additionally, we use a simplified eye model This is still a room for improvement in terms of visually appearance and realism like proposed in ~\cite{berard2016lightweight, berard2014high, berard2019practical}.

%Please note that this work does not attempt to accurately model human eyes~\cite{berard2016lightweight, berard2014high, berard2019practical}.% We instead opt to develop a static and simplified eye structure which provides pixel perfect groundtruth annotations for developing gaze estimation pipelines.

%which provides pixel perfect groundtruth annotations for developing gaze estimation pipelines.

\section{Conclusion}
This paper presents a novel synthetic eye image generation platform that provides several improvements over existing work to support the development and evaluation of eye-tracking algorithms. This platform is used to render synthetic datasets, S-NVGaze and S-OpenEDS, reflecting the spatial arrangement of eye cameras in two publicly available datasets, NVGaze and OpenEDS. We demonstrate that networks trained solely with our synthetic images have the ability to generalize to unseen eye images. We also conclude that the spatial arrangement of eye camera does not contribute as heavily as the variation in eye image appearance. Images rendered from our dataset and converged models are made publicly available to aid researchers in developing novel gaze tracking solutions.

\begin{acks}
We thank Dr. Errol Wood for providing the iris RGB textures from the Unity Eyes\cite{Wood2015RenderingEstimation} and SynthEyes~\cite{Wood2016LearningImages}. We thank Professor John Daugman\cite{Daugman2009HowWorks} for providing infrared iris textures. 
% We would also like to thank the Golisano College of Computing and Information Sciences, RIT and Research Computing, RIT~\cite{https://doi.org/10.34788/0s3g-qd15} for providing us with adequete computing resources.
\end{acks}

\newpage
% The next two lines define the bibliography style to be used, and the bibliography file.
\bibliographystyle{ACM-Reference-Format}
\bibliography{AutoUp, refs}

\end{document}